%% file: LaTeX/anonymous-submission-latex-2026.tex
\documentclass[letterpaper]{article} 
\usepackage[]{aaai2026}  
\usepackage{times}  
\usepackage{helvet}  
\usepackage{courier}  
\usepackage[hyphens]{url}  
\usepackage{graphicx} 
\usepackage{natbib}  
\usepackage{caption} 
\usepackage{algorithm}
\usepackage{algorithmic}

\usepackage{xcolor}

\usepackage{newfloat}
\usepackage{listings}

\usepackage{subcaption}

\usepackage{tcolorbox}
\tcbuselibrary{listings, breakable}
\usepackage{booktabs}
\usepackage{adjustbox}
\usepackage{makecell}

\usepackage{macros}

\DeclareCaptionStyle{ruled}{labelfont=normalfont,labelsep=colon,strut=off} 
\floatstyle{ruled}
\newfloat{listing}{tb}{lst}{}
\floatname{listing}{Listing}
\title{Benchmarking Agents in Insurance Underwriting Environments}

\author{
Amanda Dsouza,
Ramya Ramakrishnan,
Charles Dickens, \\
Bhavishya Pohani,
Christopher M Glaze
}

\affiliations{
Snorkel AI, Redwood City, CA, USA \vspace{5pt}
}

\newcommand{\method}{\textsc{Underwrite}}
\newcommand{\methodSPACE}{\textsc{Underwrite }}

\begin{document} 
\maketitle

\begin{abstract}

As AI agents integrate into enterprise applications, their evaluation demands benchmarks that reflect the complexity of real-world operations. Instead, existing benchmarks overemphasize open-domains such as code, use narrow accuracy metrics, and lack authentic complexity. We present \method, an expert-first, multi-turn insurance underwriting benchmark designed in close collaboration with domain experts to capture real-world enterprise challenges. \methodSPACE introduces critical realism factors often absent in current benchmarks: proprietary business knowledge, noisy tool interfaces, and imperfect simulated users requiring careful information gathering. Evaluating 13 frontier models, we uncover significant gaps between research lab performance and enterprise readiness: the most accurate models are not the most efficient, models hallucinate domain knowledge despite tool access, and pass\textasciicircum k results show a 20\% drop in performance. The results from \methodSPACE demonstrate that expert involvement in benchmark design is essential for realistic agent evaluation, common agentic frameworks exhibit brittleness that skews performance reporting, and hallucination detection in specialized domains demands compositional approaches. Our work provides insights for developing benchmarks that better align with enterprise deployment requirements.

\end{abstract}


\input{sections/introduction}

\input{sections/benchmark}

\input{sections/expert}

\input{sections/evaluation}

\input{sections/related_work}

\input{sections/conclusion}

\section{Ethical Statement}
This research did not involve personal data or sensitive business information. Instead, the insurance underwriting tasks involved known companies and domain-specific knowledge. The simulated environments, tasks, and data used in the benchmark were co-designed with experts to reflect realistic enterprise workflows while preserving confidentiality and excluding any proprietary or identifiable data.


\begin{quote}
\begin{small}
\bibliography{LaTeX/aaai2026}
\end{small}
\end{quote}

\clearpage
\appendix


\input{LaTeX/appendix/introduction}
\input{LaTeX/appendix/tasks}
\input{LaTeX/appendix/graph}
\input{LaTeX/appendix/hallucination_detection}
\input{LaTeX/appendix/extended_analysis}

\end{document}

%% file: sections/introduction.tex
\section{Introduction}

AI agents are increasingly being integrated into enterprise applications, characterized by proprietary business knowledge, imperfect information systems, and noisy user interactions. We, therefore, need agentic benchmarks that incorporate these complexities. Existing environments focus primarily on open-domain tasks such as coding or web reasoning, as well as emphasize narrow accuracy metrics while neglecting the  complexity of real-world operations. This mismatch has created a growing gap between academic lab performance and practical deployment. Recent meta-analyses have highlighted these limitations: \citep{zhu2025establishingbestpracticesbuilding, kapoor:tmlr25, kapoor:arxiv25}.

\begin{itemize}
    \item \textbf{Simplified or Unrealistic Environments} Many benchmarks focus on single-turn, general coding and web-interaction tasks that do not capture the complexities of enterprise workflows \cite{zhuo2024bigcodebench,yu2024humaneval}. As a result, progress measured on such benchmarks gives a misleading signal of readiness for enterprise deployment. 
    \item \textbf{Overemphasis on Accuracy} Many agentic benchmarks measure success solely through task completion or accuracy without holistic evaluation of agent behaviors that may have real-world consequences \cite{liu2023agentbench}. Because of this, agents may engage in inefficient or dangerous behaviors.
    \item \textbf{Poor Reproducibility} Many benchmarks depend on live or changing web resources or unreliable evaluators \cite{xu2024theagentcompany}. Recent work partially addresses reproducibility through self-hosted or standardized environments and fixed evaluation harnesses \cite{zhou2023webarena,chezelles2024browsergym}.
    \item \textbf{Lack of Expert Input in Environment Design} Domain experts rarely contribute to designing the environment, tasks or evaluation protocol, leading to inaccurate conclusions regarding applicability.
\end{itemize}
Thus, when an enterprise team attempts to adapt off-the-shelf agents to real workflows, unexpected failure modes emerge, like hallucinations, tool use errors, or inconsistent outputs \citep{kapoor:tmlr25, kapoor:arxiv25, cemri:arxiv25, wang2025odysseybenchevaluatingllmagents}.

To address these challenges, we introduce \method\,, an expert-first benchmark for commercial insurance underwriting. Designed in close collaboration with domain experts, \method\ simulates a realistic underwriting system that requires multi-turn reasoning, tool orchestration, and interaction with imperfect simulated users. \methodSPACE captures the realism of a real underwriting environment by incorporating proprietary business knowledge, ambiguous data, and noisy interfaces, all of which expose frontier model failure modes.


We evaluated 13 frontier models using \method\ and found that models fall short on several critical enterprise dimensions, leading to a few key insights. For example, the most accurate models are not necessarily the most efficient. Models often hallucinate domain-specific information even with tool/data access. The models show inconsistencies with the pass\textasciicircum k evaluation, revealing up to a 20\% degradation in success under realistic constraints. 

Our main contributions are as follows:
\begin{itemize}
\item \textbf{\method: An Expert-First Benchmark}
We present \method, a multi-turn, tool-augmented insurance underwriting environment that embodies proprietary knowledge, noisy tool interfaces, and imperfect user dynamics — factors central to complex enterprise systems but absent in current benchmarks.

\item \textbf{Benchmark Design Principles for Enterprise Evaluation}
We demonstrate that expert-led co-design, compositional methods of hallucination detection, and framework-robust  implementation are essential for measuring true enterprise readiness. These principles generalize beyond insurance to guide the next generation of realistic, domain-grounded benchmarks.

\item \textbf{Empirical Insights from 13 Frontier Models}  
We evaluate 13 leading models and uncover systematic gaps in accuracy, efficiency, and behavioral robustness. First, the most accurate models were not the most reliable or efficient in tool use. Second, hallucination patterns persist even with full access to domain tools, with higher rates in some small frontier model variants (GPT-5-mini has up to 19\% of completed traces have hallucinations). Finally, many frontier models are unreliable, as pass\textasciicircum k results show a drop of 20\% on answer correctness.

\end{itemize}

%% file: sections/benchmark.tex
\section{The \methodSPACE Environment}  
Here we describe our approach for constructing a realistic enterprise agent environment: defining the system/product, data and tool resources, task definition, rewards and user simulation. We considered the following desiderata in our development of \method:

\begin{itemize}
    \item \textbf{P1: Noisy Data Resources}: Real enterprises operate with imperfect data lakes, legacy systems and sometimes occasionally poor documentation.
    \item \textbf{P2: Realistic tasks and system}: We sought tasks and scenarios relevant to a real domain to focus benchmarking for complexity and skills of interest to developers in similar real-world domains.
    \item \textbf{P3: Multi-turn Interactions}: Effective multi-turn interactions remain a challenge with sizable differences in model performance when those models need to respond based on conversational history. This is also apparent in the multi-turn model scores on the Berkeley Function Calling Leaderboard \citep{patil2025the}.
    \item \textbf{P4: Challenging Reasoning Processes}: We focused on creating tasks that required a combination of reasoning, tool call chains and planning ahead of executing tool calls.
\end{itemize}

\begin{figure}[t]
\centering
\includegraphics[width=0.9\columnwidth]{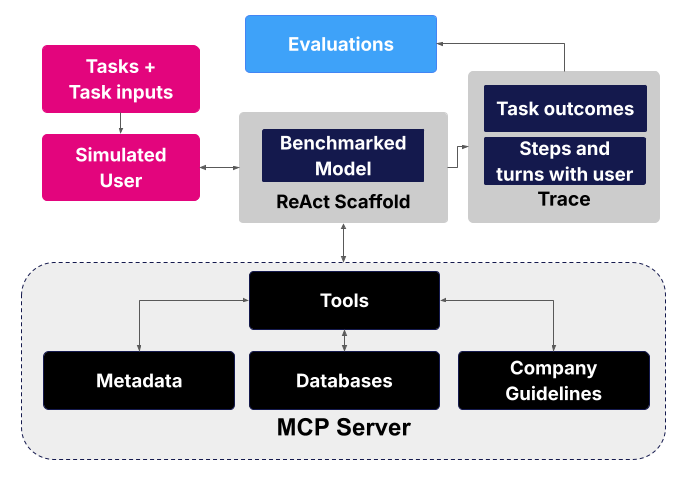} 
\caption{Architecture of the system}
\label{architecture}
\end{figure}

\subsection{Underwriting System}

We developed a system representative of the workflows in which an AI copilot would be engaged at a commercial property and casualty insurance business. The copilot's goal is to assist underwriters with decision-making about insurance applications. The environment simulates a backend that a real enterprise might build for an internal-facing product for the associated workflows. In collaboration with an expert network of Chartered Property Casualty Underwriters (CPCUs), we then developed tasks that would be the most common for a product similar to this. We aimed for an average of 3-7 steps of required reasoning and tool use, with a total of 10-20 conversational turns.

\subsection{Data and Tool Resources}
\subsubsection{Data}
We developed several key components of the system, including a SQLite database, metadata about data sources, and free-text guidelines on how the insurance company underwrites business with business rules and custom underwriting logic specific to the fictional company that would be considered proprietary at a real company. 

The database is comprised of nine tables, including -- versions of the North American Industry Classification System (NAICS) from different years, tables for qualifying a company as a small business and determining ``appetite'' matrix -- which the AI copilot could use to determine whether an applicant could qualify for specific insurance products. In addition, all resources were necessary for solving some of the more complex tasks.

Additionally, we developed metadata with descriptions of the tables to challenge models to read metadata separately before writing queries. The guidelines contained complex domain-specific logic that challenged models to read in tables via SQL queries and make decisions about further information that would need to be gathered, in addition to final decisions about task solutions. We thus incorporated environment noise with the interaction of complex logic with multiple table versions requiring fine-grained distinctions via metadata and tool descriptions.

\subsubsection{Tools}
We exposed the data resources through tools -- to view guidelines, access the database with read-only SQL queries, and tools to access metadata about the other tools. Importantly, several tools provided redundant information, challenging models to determine the appropriate tool based on information in tool descriptions. Tool descriptions for models had just enough information to enable appropriate use without any pointers as to how to distinguish similar tools from each other, another source of noise. We developed a Model Context Protocol (MCP) server to enable access to the tools and data.

\subsection{Task Definitions}
\subsubsection{Seed Task Types}
We developed six seed task types in consultation with our domain expert network, common to what underwriters would need to do in evaluating an application or communicating with the applicant directly. Task types included ``in-appetite'' determination (screening the applicant), making qualification decisions, and deciding policy limits and deductibles. 

Conversations were initiated with user requests based on one of the task types. Requests were varied using a frontier model in a way that would challenge the agent to ask follow-up questions. Consequently, we obtained task diversity through a combination of task type, user follow-up, and application characteristics, which we describe next.

\subsubsection{Task Diversity}
We varied the inputs to tasks in order to maximize task diversity and complexity. We did this by synthesizing 3000 fictional (company) applicant profiles with GPT-4.1. We ensured diversity by sampling NAICS codes with known characteristics, such as geography and revenue, desired insurance coverage and company operations. We leveraged publicly available statistics to constrain GPT-4.1 to produce realistic company profiles. 

\subsubsection{Task Complexity}
We varied task complexity by creating company profiles (and subsequent applications) that were easy or difficult to make decisions on. In the latter case, task solutions typically required longer compositions of tool calls, fine-grained distinctions of table types common in underwriting, and follow-up questions to the user. 

\subsection{Agent and User Models}
\subsubsection{Agent Framework} 
We implemented copilot (function-calling) models as ReAct agents \citep{yao2023reactsynergizingreasoningacting} in LangGraph \citep{langgraph2024} with the copilot model and simulated underwriter (the user) as separate nodes that could interact for a given number of turns. 

We included a third node in the system graph to detect errors (inappropriate responses, code responses and other indicators of errors linked with the scaffolding) in agent responses. If these indicators were detected, conversations were terminated (otherwise the simulated user would attempt to help the agent). 

We developed tools and data to be accessible via a Model Context Protocol (MCP) server \citep{mcp2024}. 

\subsubsection{Simulated User}
\methodSPACE is a multi-turn environment. We used GPT-4.1 to simulate a human underwriting user interacting with the AI copilot, with a system prompt that included instructions about the application, as well as explicit direction to respond to the agent with no more than two pieces of information about the applicant. Without this latter direction, the simulated user was prone to giving all information at once; one of the goals of th benchmark is to evaluate how effective models are at asking the right questions to solve tasks \cite{mazzaccara2024learningaskinformativequestions, zhang2024probingmultiturnplanningcapabilities, hutson2025guessinggamemeasuringinformativenessopenended}, instead of simply providing correct answers.


\subsection{Rewards}
Each task was evaluated on the following criteria:

\textit{Correctness of final outcome}: Correctness is based on reference answers created with domain rules. An LLM-as-a-Judge (GPT-4.1-mini) was used to compare the outcomes generated by the agent against the ground truth outcomes, to yield a binary reward, along with a rationale. We developed the LLM-as-a-Judge to achieve $>95\%$ agreement on manual annotations in held-out samples of 100 traces. Traces were annotated by a single expert annotator who was focused on comparing reference answers with generated answers.

\textit{Tool use errors}: Errors were determined with a rule-based function that yielded $1$ if Python exceptions were found in the agent's trace, and $0$ otherwise.

\textit{Uncertain answers}: In the graph, simulated users were instructed to answer ``I do not know" if they were asked a question they could not answer. A regular expression-based function searched for this phrase in an agent's trace to estimate the frequency of inappropriate questions. Consequently, this reward was upper bounded by the number of user turns in the interaction. Uncertain answers are used to determine the effectiveness of agents at asking relevant questions to the user.

%% file: sections/expert.tex
\section{Key Findings from Developing \method}
In this section, we summarize the key insights that emerged during the design and development of \method, focusing on realism in benchmark construction, mitigation of model scaffold errors, and evaluation strategies to shape the final benchmark.

\subsection{Expert Involvement in Benchmark Design is Key to Improving Realism}

\begin{figure}[t]
\centering
\includegraphics[width=0.9\columnwidth]{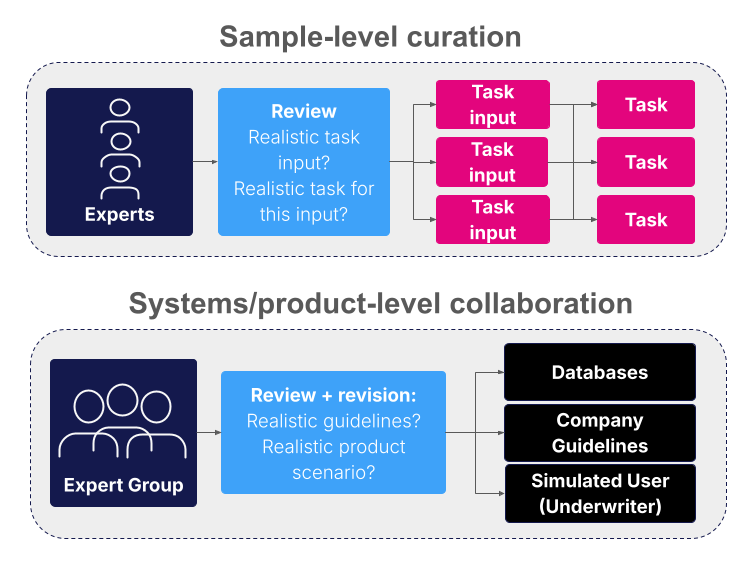} 
\caption{Representation of our expert engagement process: As in other approaches to benchmark development, experts reviewed and curated individual tasks (top diagram). However, they also collaborated as a group in developing the system more holistically as a representation of a product backend, with realistic business rules and tables (bottom diagram).}
\label{experts}
\end{figure}

Prior work on benchmarks has typically involved experts for developing \textit{individual} tasks \citep{patwardhan2025gdpvalevaluatingaimodel}, \citep{moteki2025fieldworkarenaagenticaibenchmark}. Instead, we sought an engagement model that allowed us to engage with the domain experts on the overall system, giving us feedback on how realistic the benchmark is for a product that might be deployed in an insurance company's ecosystem. 

\subsubsection{Engaging Experts to Ensure Realism of Benchmark Components}
We asked CPCUs to rate the realism of both tasks and task inputs (the fictional applicants) separately. We did so by producing traces for each task with frontier models and engaging CPCUs through an annotation interface. Interestingly, during this process, CPCUs pointed out that the same task would be relevant to some applicants but not others, suggesting a dependency we had not anticipated. We thus curated tasks for a combination of task type, task input characteristics (applicants) and the combination of each for each task, with an initial acceptance rate of 55\% after the first round of feedback and a final acceptance rate of 88\% after revising our synthetic generation process with that feedback. The final task set in our benchmark comes from only those that were accepted.

Additionally, during the annotation process, CPCUs also vetted the realism of the underwriter, by observing traces. We then developed the simulated underwriter until we achieved an acceptance rate $> 90\%$ across traces.

\subsubsection{Engaging Experts to Review and Revise the Overall System}
We engaged the CPCU network for feedback and revisions of the entire system in two sessions: one freeform, 15-minute session with feedback on the entire scenario and task set they had reviewed, and one deeper 30-minute session in which CPCUs reviewed guidelines and data resources, suggesting changes that we then aggregated.

While relatively time-consuming from a development perspective, we found the 45 minutes total of feedback to have obviated the need for more rounds of task-level curation.

\subsection{Agentic Frameworks Exhibit Brittleness that Skews Performance Reporting}
A key technical challenge in developing \methodSPACE was disentangling model failures from framework artifacts. We carefully designed our evaluation to ensure that task failures (Table \ref{table:failure_modes}) reflected genuine model limitations rather than brittleness in the LangGraph scaffold \cite{kapoor:arxiv25}, addressing a common source of inaccurate performance reporting in existing benchmarks \citep{zhu2025establishingbestpracticesbuilding, kapoor:arxiv25}. This took several development cycles focused on code detection and other inappropriate responses to the user. Development work included robust methods to parse out thinking tokens from open source models that expose those, and lightweight verifiers to detect inappropriate code blocks that are occasionally expressed to users.

\subsection{Hallucination Detection is Difficult in Complex Business Domains}
During exploratory analysis, we observed a distinct error mode related to hallucinations based on pretrained domain knowledge, with frequent references to insurance products nowhere to be found in the guidelines. We found developing a more general hallucination detection metric to be difficult, with most LLM-as-Judge based approaches resulting in high recall but low precision; incorrectly labeling features of agent rationales as hallucinations (rather than, for instance, incorrect reasoning).
We thus developed a conservative hallucination-detection language model (GPT-4.1) focused on the specific hallucinations on insurance products.  

%% file: sections/evaluation.tex
\section{Evaluating Models on \method}
\label{sec:empirical_analysis}

\subsection{Experimental Details}

We evaluated various proprietary and open-source function-calling models on \method, on 300 tasks. Agents were allowed to interact with the simulated user for 50 turns. Upon task completion, or a time-out, the reward criteria were used to evaluate performance. We report pass\textasciicircum 1 as well as pass\textasciicircum k scores. We used default model parameters for all agents (including temperature, reasoning budget and token limits), observing only a small handful of API errors related to token limits during rollouts (see Table \ref{table:failure_modes}).

\subsection{Results and Analysis}
\subsubsection{The most accurate models were not the most efficient, based on \method\ reward criteria.} We evaluated 13 frontier models on all three reward criteria defined in \method. As shown in Table \ref{table:model_comparison}, frontier models had a wide range in correctness scores, from 30\% to 88\%. Among open source models, DeepSeek-V3.1 and Qwen3-Coder-480B are competitive, outperforming Gemini-2.5-Pro and ``smaller'' GPT-5 variants (GPT-5-Nano, GPT-5-Mini), among others. In addition to answer correctness, \methodSPACE also tests performance with respect to tool errors, and unknown answers (a proxy to how effective models are at eliciting information from the user). Results show that the top performing model, Claude-Sonnet-4.5 exhibits higher than average tool call errors (average per trace) and rate of uncertain user responses. In contrast, the cheaper, but faster, Claude-Haiku-4.5 has a lower answer correctness (77.3\% instead of 88.3\%) but lower average-per-trace tool error and uncertain user response rates.

\subsubsection{High-performing models were able to successfully reason internally with limited, efficient interaction with the user.} 
\begin{figure*}[ht]
    \centering
     \begin{subfigure}{0.4\textwidth}
        \centering
        \textbf{Gemini 2.5 Pro}
        \includegraphics[width=\textwidth]{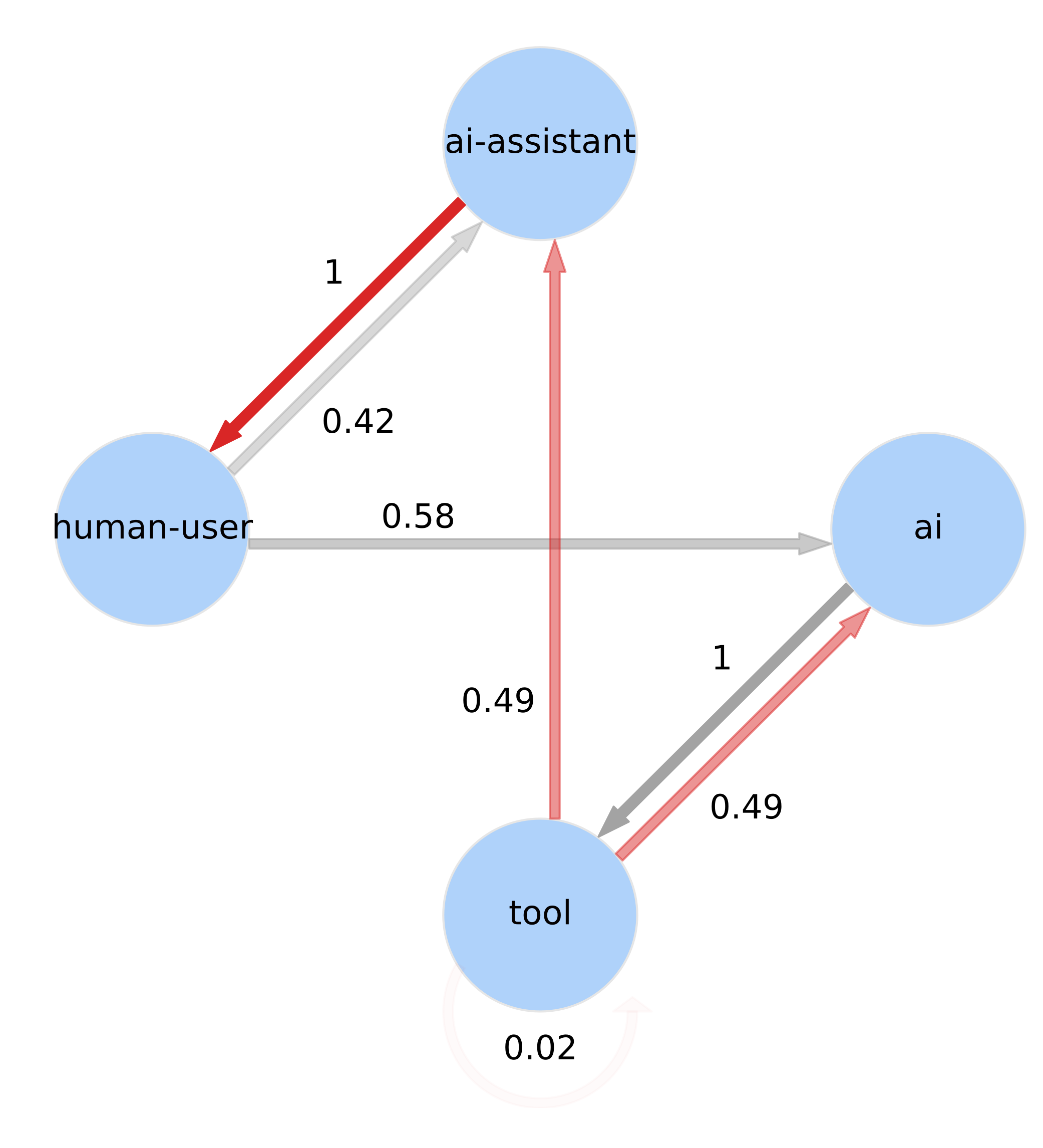}
    \end{subfigure}
    \hspace{1em}
    \begin{subfigure}{0.4\textwidth}
        \centering
        \textbf{Claude Sonnet 4.5}
        \includegraphics[width=\textwidth]{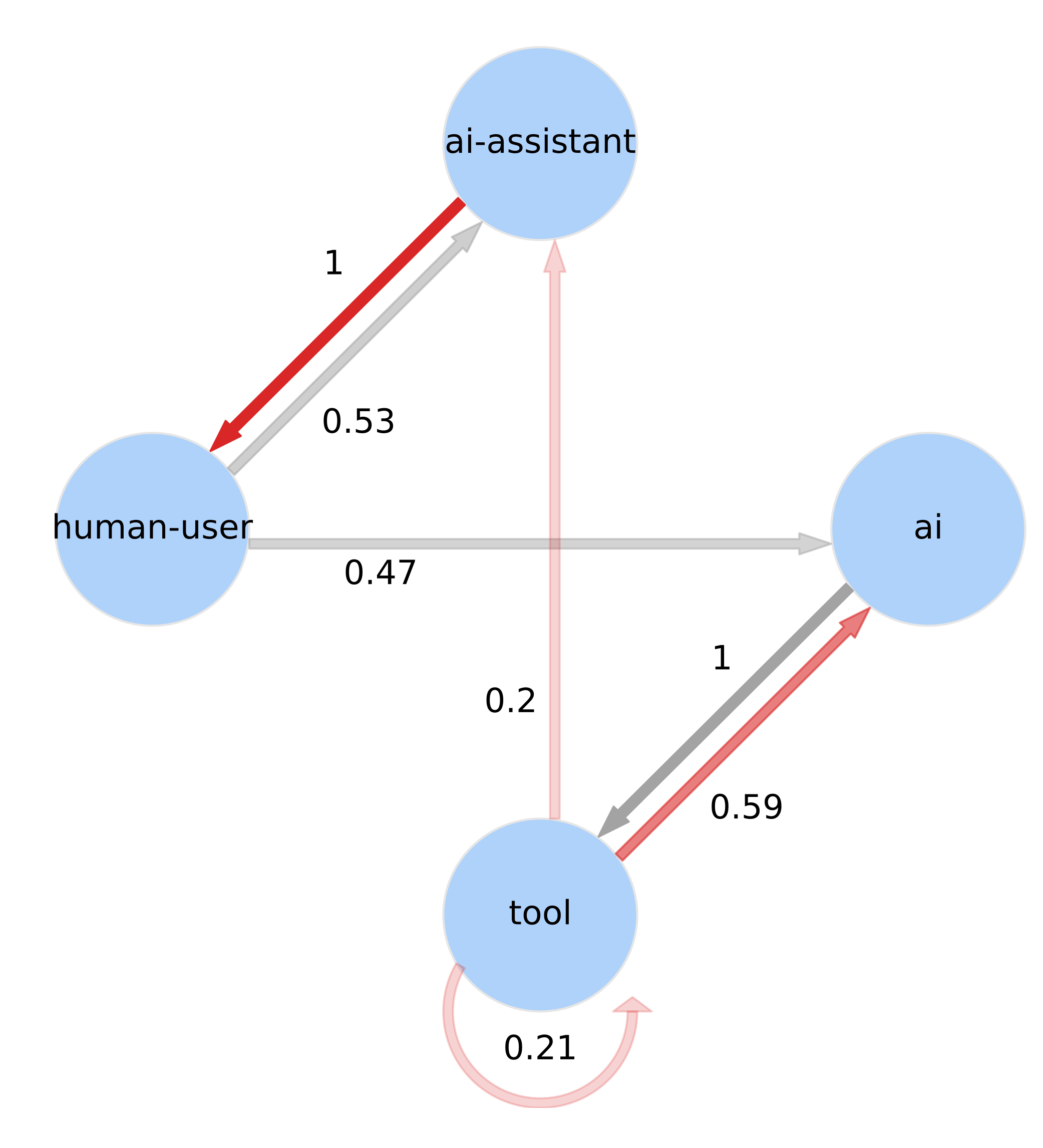}
    \end{subfigure}
    \hfill

    \caption{State transition diagrams for Gemini-2.5-Pro and Claude-Sonnet-4.5 agents. ``ai'' refers to the ReAct agent's thinking steps, and ``ai-assistant'' denotes the user-facing steps. Gemini-2.5-Pro takes more turns, sending the results of tools directly back to the user more often (tool$\rightarrow$ai-assistant and ai-assistant$\rightarrow$user transitions). This is in contrast to Claude-Sonnet-4.5, which is more apt to reason over multiple tool call results (tool$\rightarrow$tool and tool$\rightarrow$ai transitions) before turning back to the user.}
    \label{fig:transitions}
\end{figure*}

As shown in Figure \ref{fig:transitions}, we also see significant variations via state transition diagrams in the dynamics of model interactions step-by-step. For example, some of the highest performing models such as Claude-Sonnet-4.5 were able to make multiple tool calls in a turn with more internal reflection, often self-correcting tool errors. In contrast, others such as Gemini-2.5-Pro were more apt to send responses directly back to users, sometimes with inappropriate questions that elicited ``I do not know" responses from the user. We analyze some of these error modes more below. 

\begin{table*}[h]
\centering
\begin{tabular}{lccc}
\toprule
\textbf{Model} & \textbf{Answer Correctness} & \textbf{Tool Errors} & \textbf{User Cannot Answer} \\
 & \textbf{(\%)}  & \textbf{(Avg per trace)} & \textbf{(Avg per trace)} \\
\midrule
Claude Sonnet 4.5 & \textbf{90.30} & 0.520 & 0.28 \\
GPT 5 & 83.30 & 0.393 & 0.20 \\
Grok 4 & 83.30 & 0.233 & 0.33 \\
Claude Haiku 4.5 & 77.30 & 0.340 & \textbf{0.15} \\
DeepSeek V3.1 & 73.70 & 0.777 & 0.29 \\
Qwen3 Coder 480B & 73.30 & 0.343 & 0.29 \\
Gemini 2.5 Pro & 72.70 & 0.663 & 1.61 \\
GPT 5 Mini & 71.70 & 0.793 & 0.27 \\
Gemini 2.5 Flash & 62.00 & 0.513 & 1.13 \\
Kimi K2 Instruct & 56.70 & 0.803 & 0.41 \\
GPT 5 Nano & 47.00 & 0.263 & 1.53 \\
Qwen3 235B & 30.00 & \textbf{0.200} & 0.48 \\
GPT OSS 120B & 30.00 & 0.213 & 0.29 \\
\bottomrule
\end{tabular}
\caption{Overall model performance by reward criteria. ``answer correct'' denotes the proportion of tasks in which the final response to the user had a correct solution. ``tool errors'' is the average number of exceptions thrown for each task, while ``user cannot answer'' is the average number of times per task the user responded with uncertainty to the agent.}
\label{table:model_comparison}
\end{table*}


\begin{table*}[h]
\centering
\begin{tabular}{lccccc}
\toprule
\textbf{Model} & \textbf{Incomplete} & \textbf{Max Steps} & \textbf{Empty}  & \textbf{Code}  & \textbf{API Error} \\
\midrule
GPT OSS 120B & 24.30 & 0.00 & 16.70 & 11.00 & 0.00 \\
Kimi K2 Instruct & 24.00 & 0.00 & 0.00 & 25.30 & 0.00 \\
Gemini 2.5 Flash & 14.30 & 0.00 & 8.30 & 6.30 & 0.00 \\
Grok 4 & 9.30 & 0.00 & 9.30 & 0.00 & 0.00 \\
Qwen3 235B & 5.70 & 0.00 & 0.00 & 8.00 & 0.70 \\
DeepSeek V3.1 & 1.30 & 0.00 & 0.70 & 0.70 & 0.00 \\
GPT 5 Nano & 1.00 & 1.00 & 0.00 & 0.00 & 0.00 \\
Claude Haiku 4.5 & 0.30 & 0.00 & 0.30 & 0.00 & 0.00 \\
Claude Sonnet 4.5 & 0.30 & 0.00 & 0.30 & 0.00 & 0.00 \\
Gemini 2.5 Pro & 0.30 & 0.00 & 0.30 & 0.00 & 0.00 \\
Qwen3 Coder 480B & 0.00 & 0.00 & 0.00 & 0.00 & 0.00 \\
GPT 5 & 0.00 & 0.00 & 0.00 & 0.00 & 0.00 \\
GPT 5 Mini & 0.00 & 0.00 & 0.00 & 0.00 & 0.00 \\
\bottomrule
\end{tabular}
\caption{Incomplete task rates by model. 
The first column is the overall incomplete rate, while subsequent columns denote root causes of the failures (root causes are not mutually exclusive, therefore do not sum to the overall incomplete rate).}
\label{table:failure_modes}
\end{table*}

\begin{figure}
    \centering
    \includegraphics[width=1.0\linewidth]{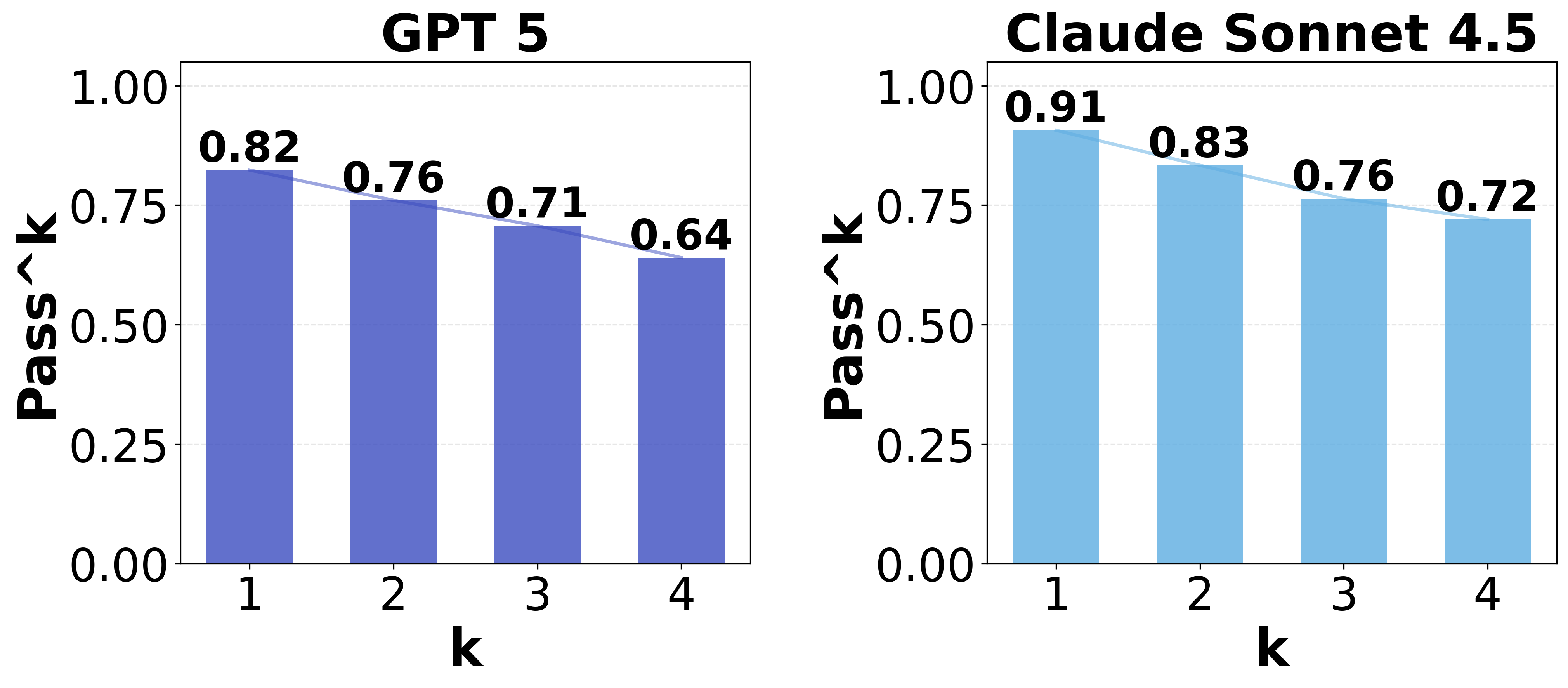}
    \caption{pass\textasciicircum k measured on Answer Correctness across all task types for GPT-5 and Claude-Sonnet-4.5.}
    \label{fig:pass_at_k}
\end{figure}



\begin{figure}[t]
\centering
\includegraphics[width=0.9\columnwidth]{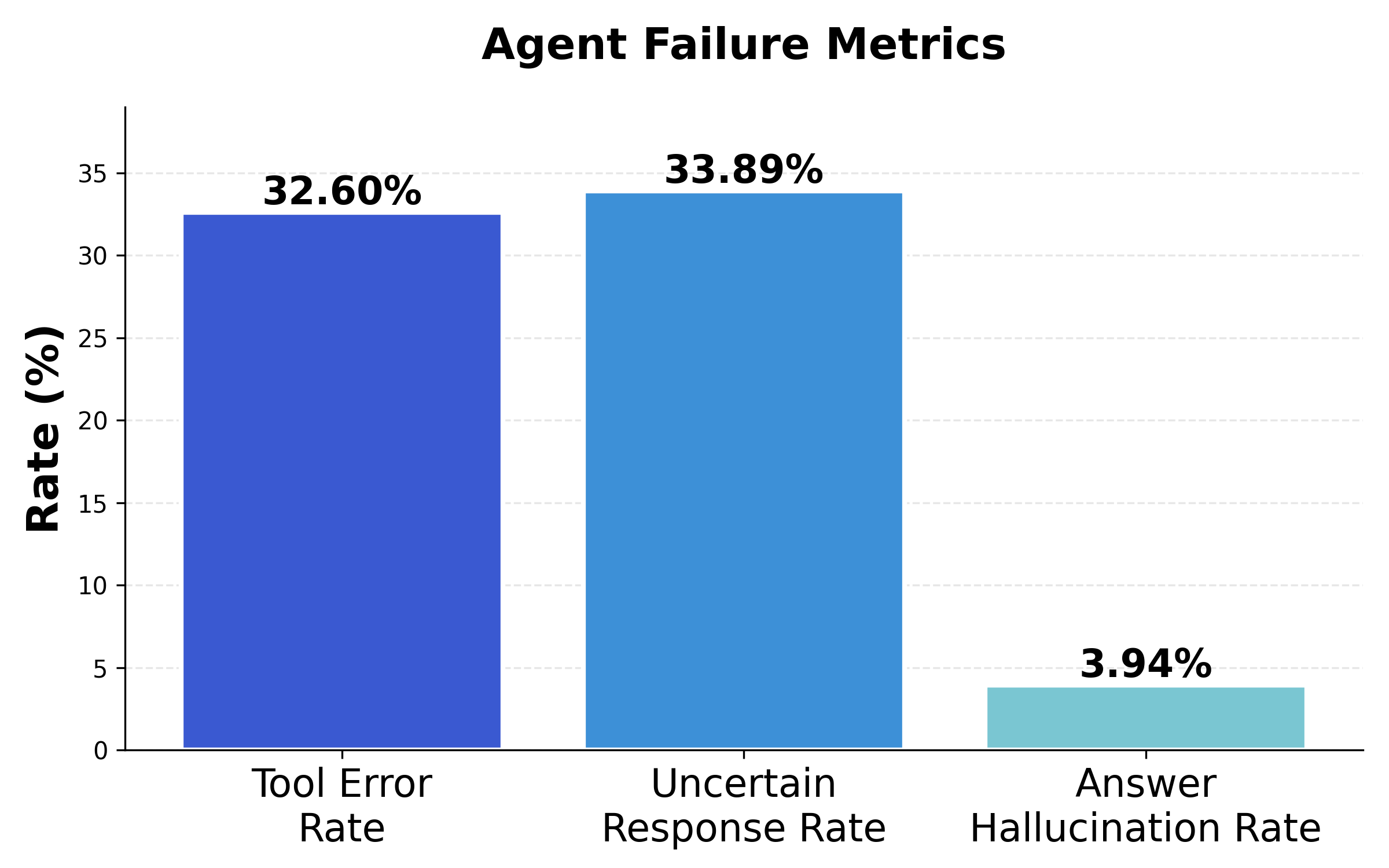} 
\caption{Failure rates of agents, aggregated over all completed traces, and models. Tool error rate measures the \% of traces with at least one tool error. Similarly, uncertain response rate and answer hallucination rate measures the \% of traces with at least one uncertain response, and one hallucinated answer, respectively.}
\label{failure_plot}
\end{figure}

\begin{figure}[t]
\centering
\includegraphics[width=0.9\columnwidth]{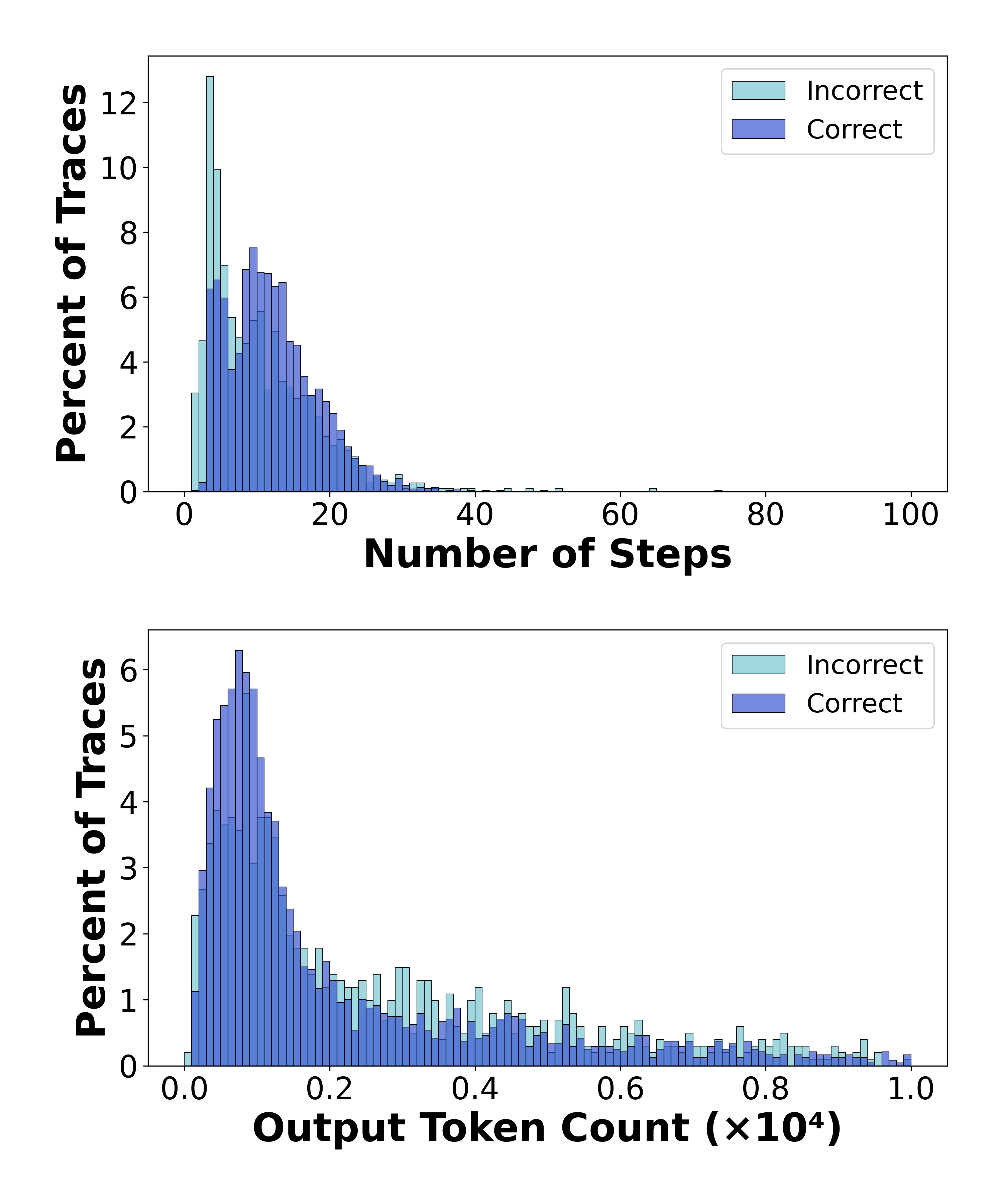} 
\caption{Completed traces of incorrect outcomes, across all models, tend to use fewer steps, but a higher number of tokens, indicating more verbose responses, as compared to those of correct outcomes.}
\label{histogram}
\end{figure}

\begin{figure}[ht]
\centering
\includegraphics[width=1.0\columnwidth]{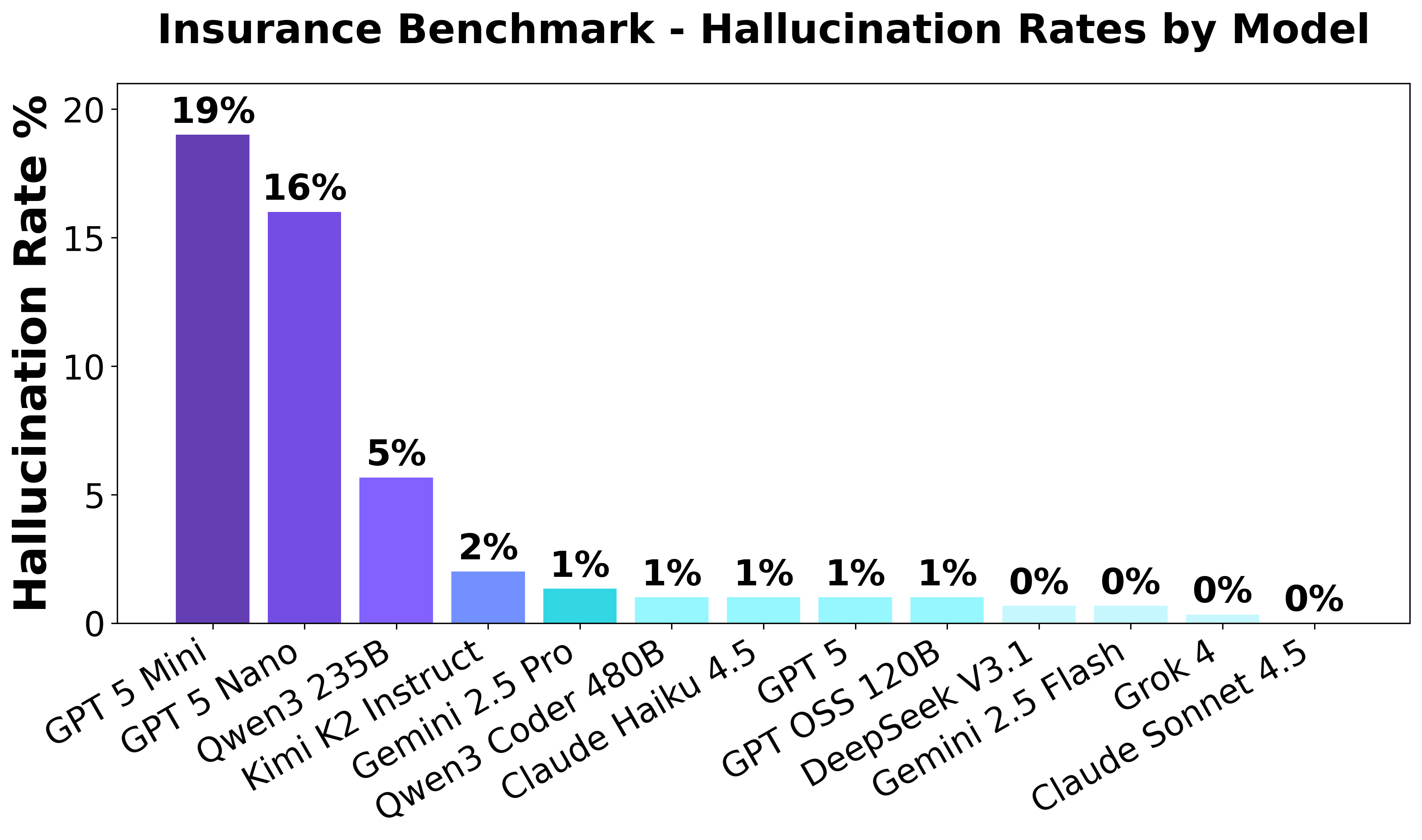} 
\caption{Hallucination rates of models measured over all completed traces. Answer hallucination rate measures the \% of traces with at least one  hallucinated answer.}
\label{hallucination}
\end{figure}

\subsubsection{Models hallucinate based on pretrained domain knowledge.} We observed that some of the frontier models had high hallucination rates, as shown in Figure \ref{hallucination}. In the task involving insurance product recommendations, the smaller OpenAI models (GPT-5-Mini, GPT-5-Nano) hallucinated several products not in the guidelines 58-66\% of the time (Figure \ref{fig:hallucination_top_3} of Appendix). These hallucinations not only led to misleading answers, but also to misleading questions to the underwriter—probing for information that was ultimately irrelevant to the task. Additionally, we found that product hallucination rates were moderately correlated, in the negative direction, with accuracy, as shown in Table \ref{correlation} in Appendix.

More details about our hallucination detection methodology, including the evaluation setup and the system prompt used to isolate errors, are provided in the Appendix.

\subsubsection{Completion failures and pass\textasciicircum k metrics reveal critical reliability gaps across models.}
Evaluating reliability is a key consideration for deploying models in production, on our simulated \methodSPACE
system. We assess the reliability of agents by measuring completion failures, as shown in Table \ref{table:failure_modes}. Among the evaluated models, GPT-OSS-120B exhibits the highest rate of empty completions, while Kimi-K2-Instruct most frequently produces non-parseable responses. Both behaviors contribute to forceful termination of responses, resulting in incomplete outputs. 

We also analyzed model reliability in enterprise settings through pass\textasciicircum k metrics, as is done in prior work \cite{yao2024taubenchbenchmarktoolagentuserinteraction,barres:arxiv25}. We computed the pass\textasciicircum k metric by running each model through the evaluation framework up to $k$ times ($k \in \{1,\dots,4\}$). As shown in Figure \ref{fig:pass_at_k}, we observed a roughly 20\% drop in correctness when testing up to 4 trials. This highlights the importance of building systems that are consistent and reliable, not just successful on average.

\subsubsection{Number of tool use errors has weak correlation with overall performance.} 
Across models, including top performers, agents made at least one tool call error in 32\% of the conversations (Figure \ref{failure_plot}).
These errors occurred despite access to metadata describing the tool functionality and correct usage. 
Surprisingly, we found a weak correlation between tool user error rates and overall performance, as shown by the reported tool error rates in Table \ref{table:model_comparison}.
The highest correlation observed was in Product Recommendations tasks, which had a low-to-moderate Pearson correlation of $0.256$ (Table \ref{correlation} in Appendix).
In all other cases, even the three most accurate models made tool call errors in $20-40\%$ of the conversations (Figure \ref{failure_plot_top} in Appendix), often going back after basic tool call errors to retrieve metadata and redo the tool call correctly.
We did, however, find a moderate to strong positive correlation between correctness and the recovery rate defined as the rate of tool errors with a subsequent successful call to the same tool (Table \ref{correlation} in Appendix).
This result indicates that tool call errors are relatively common across all models, and part of what differentiates higher-performing agents is self-correction capabilities.

\subsubsection{Number of agent steps and token counts are moderately correlated with overall performance.} 
As shown in Figure \ref{histogram}, on average, completed agent traces across all models, with incorrect outcomes, have fewer steps but a higher number of tokens.
For instance, in one task, GPT-5-Nano generated over $7$k output tokens across only $3$ steps and did not make any tool calls. 
Instead, GPT-5-Nano relied on reasoning and asking questions to the simulated user to reach an incorrect final answer.
On the same task, Claude-Haiku-4.5 generated under $400$ output tokens across $4$ total steps, including 1 successful tool call to reach a correct final answer.
The results in Figure \ref{histogram} and the example highlight an observed pattern of verbose responses with fewer turns and hence fewer successful tool calls leading to incorrect answers. 


\subsubsection{Model accuracy is negatively correlated with degree of proprietary information in answers.} 

As discussed in previous sections, one area we engaged with experts the most was in developing proprietary information that was fictional but plausible, challenging models to reason with the available resources instead of relying on pretrained knowledge stored in model weights. We directly investigated this in the results by rating each task (post-hoc) with respect to how ``surprising'' reference answers were to the two most accurate frontier models (GPT-5 and Claude-Sonnet-4.5). Specifically, we probed both models by providing them, for each task, with the full inputs required to solve the task along with the reference solution, ablating access to the resources. We then asked each model to rate on an ordinal scale of 1-5 how surprising the reference solutions are. The two models were only modestly correlated in their ratings Spearman's rank correlation coefficient of r=0.34), so we averaged ratings for an aggregate ``surprise" score. 

\begin{figure}[h]
    \centering
    \includegraphics[width=0.9\columnwidth]{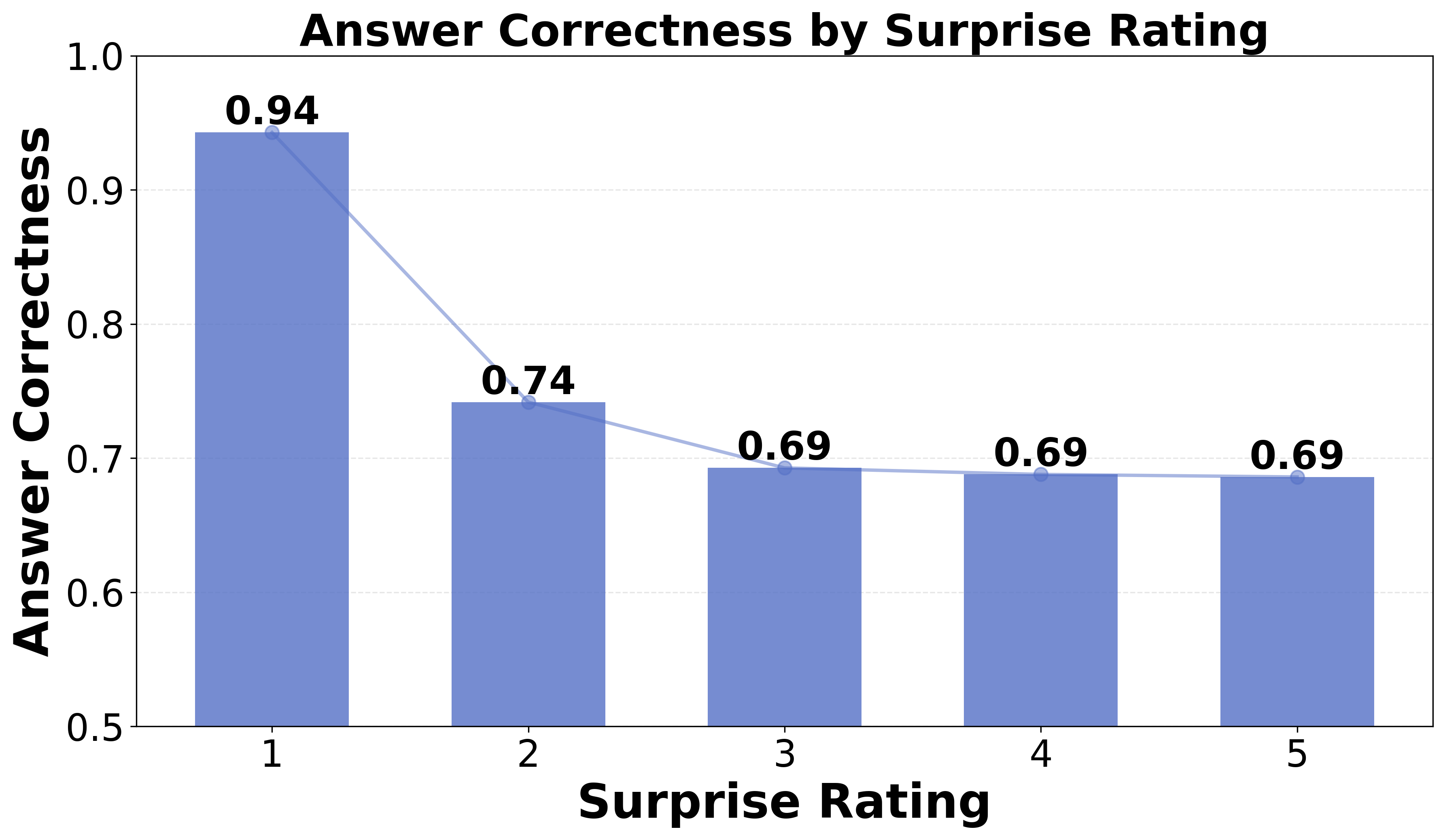}
    \caption{Model accuracy by reference answer ``surprise". Results indicate very high model accuracy for the least surprising reference answers, with a sharp drop as surprise increases. The data overall suggest that a large driver of model inaccuracy stems from the proprietary knowledge developed with experts.}
    \label{fig:surprise}
\end{figure}

Indeed, across all models, accuracy sharply dropped with surprise (Figure \ref{fig:surprise}), suggesting that proprietary information was also a strong driver of model errors. While two of the models evaluated were also used to probe surprise, we actually found those to have the weakest correlation, with the smaller models showing the strongest. For example, Claude-Haiku-4.5 was close to $100\%$ accurate on tasks with the lowest surprise rating vs $66\%$ accurate on tasks with the highest, while Claude-Sonnet-4.5 was $92\%$ accurate on low-surprise tasks and $88\%$ accurate on high-surprise tasks.

%% file: sections/related_work.tex
\section{Related Work}
In this section, we review related literature on domain specific agentic benchmarks, expert involvement in benchmark design, and best practices for building benchmarks.
See \citep{zhang:arxiv25} for a comprehensive survey on agentic reinforcement learning covering environments and benchmarks and see \citep{yehudai:arxiv25} for a focused survey on evaluation methodologies for agents.


\textbf{Domain Specialized Benchmarks}
Our work is most related to benchmarks evaluating agents in real world tasks requiring expert domain knowledge and skills.
Benchmarks have been developed to evaluate agents in the domains of machine learning research and engineering \citep{starace:icml25, chan:iclr25},
software engineering \citep{jimenez:iclr24},
biomedical \citep{xu:arxiv25}
airline, retail, and telecommunications \citep{yao:iclr25, barres:arxiv25}.
\method\, contributes to this high impact direction of research with the first benchmark designed for commercial insurance underwriting.

\textbf{Use of Experts in Benchmark Development}
Domain experts are vital to benchmark development to ensure evaluations accurately reflect real-world outcomes.
Experts have been typically leveraged to develop tasks requiring expertise and evaluate the agent's outputs. For instance, \citenoun{patwardhan2025gdpvalevaluatingaimodel} used experts to develop and evaluate model capabilities on high economic-value tasks. Similarly, \citenoun{xu:arxiv25} sourced tasks and data from real-world healthcare settings and validated by a panel of healthcare experts. 

An emerging use of expert input is in the environment design phase to ensure authenticity. Terminal-Bench evaluates agents in a terminal environment \citep{tbench:misc25}, where experts contribute self contained environments -- problem descriptions, docker containers, and tests for evaluation. \method\, extends this paradigm by leveraging domain experts in the environment design to capture authentic enterprise settings. This approach aligns with emerging best practices in benchmark creation to ensure environment validity and practicality.

\textbf{Agent Benchmark Standards}
Recent research has made progress towards establishing guidelines and best practices for agentic benchmarks \citep{zhu2025establishingbestpracticesbuilding, kapoor:tmlr25, kapoor:arxiv25}. 
\citep{zhu2025establishingbestpracticesbuilding} introduce the Agentic Benchmark Checklist (ABC), a framework for evaluating the the task validity, outcome validity, and transparent reporting. 
\citep{kapoor:tmlr25} emphasizes cost-aware evaluation and reproducibility protocols, advocating for benchmarks that optimize accuracy and efficiency with documented evaluation procedures.
Our learnings, while developed for the underwriting benchmark, are broadly applicable to benchmark design.

%% file: sections/conclusion.tex
\section{Conclusion}

In this work, we introduced \method, an expert-first, agentic benchmark for multi-turn insurance underwriting, that captures the complexity of enterprise settings. Through close collaboration with domain experts, \methodSPACE creates a more realistic environment, integrating proprietary knowledge, imperfect user behavior, and noisy tool interfaces. Evaluating 13 frontier models, we found several shortcomings: the most accurate models were not the most efficient, hallucinations persisted despite tool access, and pass\textasciicircum k evaluation revealed up to 20\% performance degradation. Our work demonstrates three key principles for enterprise-ready agentic benchmarks. First, expert involvement must extend beyond task creation to holistic environment co-design. Second, compositional approaches are essential for detecting hallucinations in complex domains. Third, framework-robust evaluation is necessary to isolate model capabilities from scaffold artifacts. These principles generalize beyond insurance to any specialized domain with proprietary knowledge, noisy data, and high-stakes decision-making. By prioritizing expert collaboration and realistic complexity, we can build benchmarks that truly measure enterprise readiness and bridge the gap between academic progress and practical deployment.

%% file: LaTeX/appendix/introduction.tex
\section{Appendix}
\label{appendix:intro}

%% file: LaTeX/appendix/tasks.tex
\section{Task Details}
\label{app:task_description}
\subsection{Seed Task Types}
\label{app:seed_tasks}
We describe in detail the seed task types included in \method:

\begin{itemize}
  \item Whether the type of insurance, or “line of business,” being applied for was “in appetite,” which is an initial screening to see if the company attributes are acceptable to the fictional underwriting company.
  \item What, if any, other types of insurance the underwriter (the user) should offer the applicant based on their characteristics.
  \item Whether the applicant qualifies as a “small business.”
  \item The appropriate classification of the applicant based on their operations using the NAICS schema.
  \item What types of policy limits the underwriter (the user) should offer the applicant if they are in appetite.
  \item What types of deductibles the underwriter (the user) should offer the applicant if they are in appetite.
\end{itemize}

\subsection{Inputs to Tasks}
\label{app:task_inputs}
Inputs to tasks included the following attributes:
\begin{itemize}
    \item Company name
    \item Verbose description of the applicant’s operations
    \item Number of employees
    \item Annual revenue
    \item Total annual payroll
    \item Number of automobiles operated and owned (if auto insurance was relevant)
    \item Basic description of the property (if property insurance was relevant)
    \item Location
\end{itemize}

\subsection{Task complexity example - Difficult}
\label{app:task_difficult}
To determine whether an applicant even qualifies as a small business, AI copilots had to: (1) Find the proper NAICS classification code from the 2012 version of the schema, (2) use this code to query a table from the US Small Business Administration on which feature of the business to use for qualification (number of employees vs annual revenue), (3) use the value of that threshold. Agents only had primary access to information about 2022 NAICS codes and mappings to the 2012 version via two other tables. They thus they had to enter a chain of SQL queries to determine the correct criteria and thresholds, interacting with underwriters to obtain the information they needed in the process.

\subsection{Task complexity example - Easy}
\label{app:task_easy}
To determine whether an applicant was in appetite for property insurance, agents had to (1) read the free-text underwriting guidelines, (2) determine whether the applicant was in a special cohort of class codes (related to real estate), and (3) if the applicant was, gather information about the property, classify the property construction, and using all of the above to make a final decision.

%% file: LaTeX/appendix/graph.tex
\section{Graph Details}

\begin{figure}[ht]
    \centering
    \includegraphics[width=1.0\columnwidth]{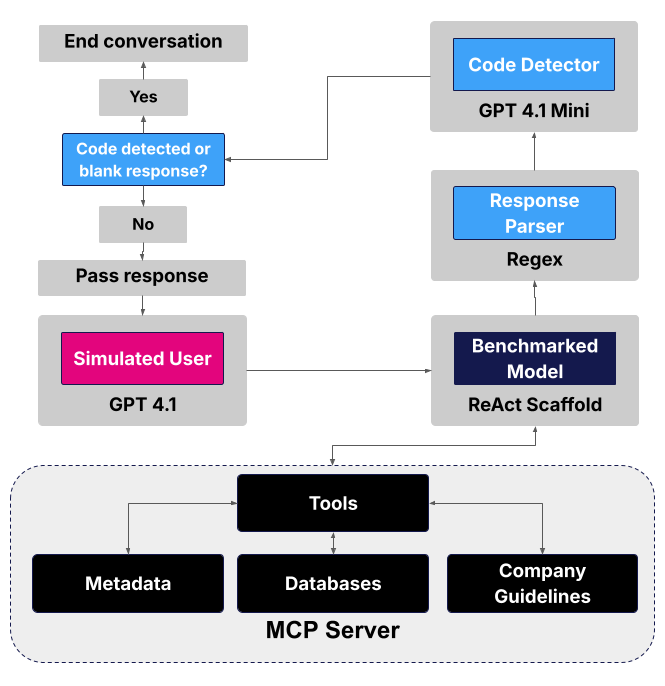}
    \caption{More granular view of graph subserving the environment.}
    \label{fig:granular_graph}
\end{figure}

Figure \ref{fig:granular_graph} presents a more granular view of the graph we built in Langgraph for the environment.

Some details:

\textbf{Response parser}: this module uses regex to parse out any text within \texttt{<s>...</s>} where \texttt{s} is any alphanumeric sequence.

\textbf{Code detector}:  a lightweight language model for detecting code with structured output and the following prompt:

\begin{figure}[ht]
\begin{tcolorbox}[colback=gray!5!white, colframe=gray!80!white, left=0.8em, right=0.8em, boxsep=0.6em, title= \textbf{Code Detector Prompt}, breakable ]
\medskip
You are an AI assistant who analyzes text and is good at analyzing text. Please tell me whether the following TEXT contains any computer code or JSON at all. Do NOT count simple references to tables or other codes such as classifications. For example, 'NAICS code' references are NOT examples of computer code.

\bigskip

TEXT:
\{response\}
\end{tcolorbox}
\label{fig:codeprompt}
\end{figure}

%% file: LaTeX/appendix/hallucination_detection.tex
\section{Hallucination Detection Details}
\label{appendix:hallucination_detection}

Here we describe in more detail how we developed our focused hallucination detection model. The goal was to increase confidence in the signal detected (minimize false positives) so that we could identify the prevalent hallucinations we observed in traces during exploratory analysis. Earlier work began with more general hallucination detection with a language model but we found very low precision.

To avoid false positives, we thus focused our language model (GPT 4.1 Mini) on (a) identifying entities and (b) distinguishing entities mentioned in the fictional guidelines from those not mentioned. We iterated on this prompt with a test bed of 50 manually annotated samples by an expert and achieved \textgreater95\% accuracy on a balanced sample of responses with and without hallucinations. The final prompt is shown below.

\begin{figure}[ht]
\begin{tcolorbox}[colback=gray!5!white, colframe=gray!80!white, left=0.8em, right=0.8em, boxsep=0.6em, title= \textbf{Hallucination Detection Prompt}, breakable ]
\vspace{5pt} 
You are an AI assistant for commercial property insurance and your task is to evaluate ANSWERS generated for underwriters at All National Insurance. 
\medskip
All National Insurance underwrites the policies for small businesses with the following lines of business (LOBs):
\begin{itemize}
\item property
\item general liability
\item auto
\item workers compensation
\item cyber
\item BOP (business owners policy)
\end{itemize}

\medskip
For the below ANSWER, please list all insurance products mentioned (if any), along with an indicator as to whether each insurance product is hallucinated (ie a part of the above list).

\medskip

Tips:
\begin{itemize}
\item IGNORE statements in ANSWER about limits, deductibles, appetite, NAICS codes and other statements specific to the applicant for insurance.
\item ONLY base your judgment on facts about insurance products at All National Insurance.
\item Focus mostly on whether any products (lobs) are hallucinated. For example, if ANSWER references products such as Inland Marine, Equipment Breakdown, Environmental, Errors and Omissions, it as a hallucination because none of those products are in the guidelines.
\end{itemize}

\medskip
ANSWER:
\{response\}
\end{tcolorbox}
\label{fig:hallucinationprompt}
\end{figure}

Language model judge outputs were structured, listing products in a list of dictionaries with product description and whether it was hallucinated. In post-hoc analysis we parsed out listed products from structured outputs and made regex-based corrections to a few common references that were listed as hallucinated that were not. For example, "commercial auto" was occasionally identified as a hallucination so we replaced those judgments with "no hallucination" indicators.

All hallucinated products were common ones in the insurance industry (and simply not offered by the fictional company, which is common), strongly suggesting that they come from pretrained domain knowledge.

%% file: LaTeX/appendix/extended_analysis.tex
\section{Extended Evaluation}
\label{app:extended_empirical_evaluation}
In this section, we expand on our empirical analysis presented in the main paper.

\subsection{Top Model Failure Modes}
\label{app:top_model_failure_modes}

\begin{figure}[ht]
\centering
\includegraphics[width=0.9\columnwidth]{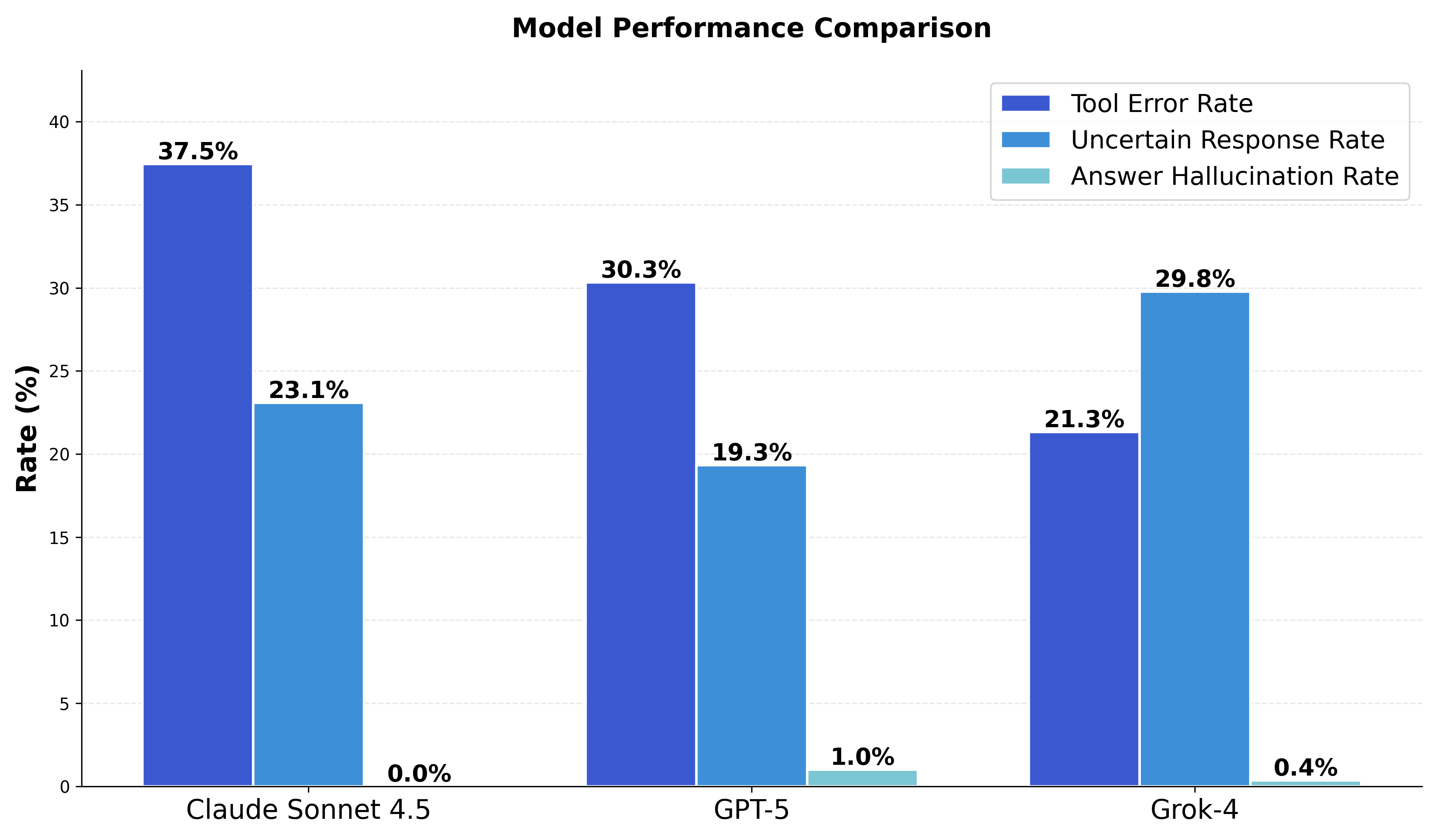} 
\caption{Failure mode rates of 3 models with the highest answer correctness, measured over all completed traces.}
\label{failure_plot_top}
\end{figure}

\begin{figure}[ht]
    \centering
    \includegraphics[width=0.9\columnwidth]{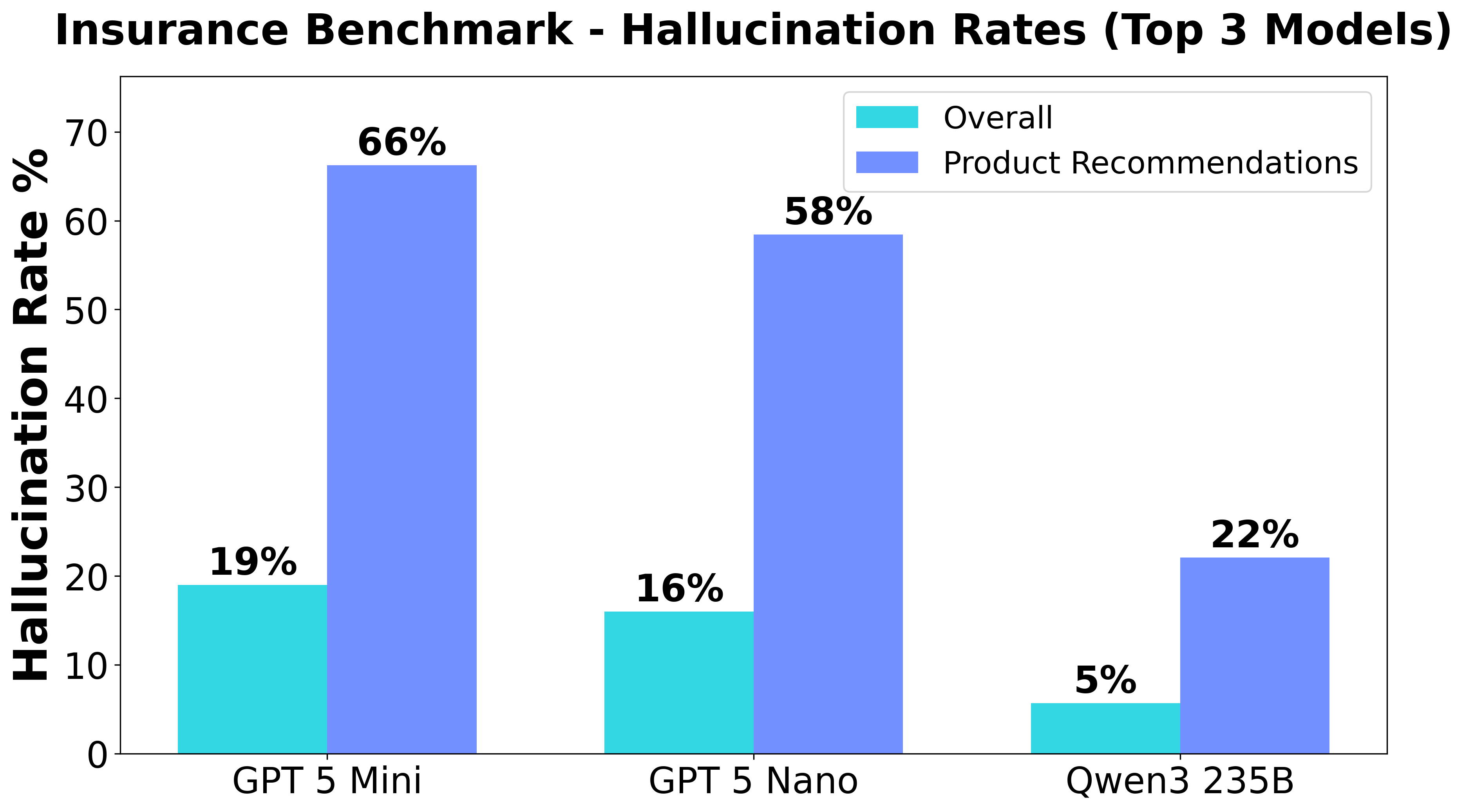}
    \caption{Models with highest hallucination rates split by task type (\textit{Overall} vs. \textit{Product Recommendations}).     The difference between overall and product-specific hallucinations highlights that models tend to hallucinate more frequently when producing product recommendations, indicating task-specific reliability gaps.}
    \label{fig:hallucination_top_3}
\end{figure}

We analyzed the rates of common failure modes in completed traces for the 3 top-performing models according to answer correctness reported in Table \ref{table:model_comparison}.
Specifically, we measured the tool error rate, uncertain response rate, and answer hallucination rate for Claud Sonnet 4.5, GPT-5, and Grok-4.
Our findings are reported in Figure \ref{failure_plot_top}.
Aligned with the result reported in main paper, even accurate models have a high tool error rate.
In the main paper, the tool error rate was reported across all traces, while Figure \ref{failure_plot_top} shows a similar pattern on completed traces.
The uncertain response rate varies across top-performing models and was substantial.
Finally, we consistently saw low answer hallucination rates.
These results all indicate a correlation between low hallucination rates and a robustness to uncertainty and tool errors with answer accuracy. 


\subsection{Correlation of Indicators with Answer Correctness}
\label{app:correlation_analysis}

\begin{table*}[ht]
\small
\centering
\begin{tabular}{lcccccc}
\toprule
\textbf{Task} & \textbf{Num Steps} & \makecell{\textbf{Token} \\ \textbf{Counts}} & \textbf{Tool Errors} & \makecell{\textbf{Uncertain}\\ \textbf{Count}} & \textbf{Hallucinations} & \makecell{\textbf{Tool Error} \\ \textbf{Recovery}} \\
\midrule
Appetite Check & 0.21 & -0.07 & 0.08 & -0.15 & 0.00 & \textbf{0.84} \\
Small Business Eligibility Check & 0.19 & 0.04 & 0.14 & -0.06 & - & \textbf{0.41} \\
Deductibles & 0.22 & -0.20 & 0.13 & -0.02 & -0.17 & \textbf{0.68} \\
Business Classification & -0.21 & -0.27 & 0.01 & \textbf{-0.29} & - & - \\
Product Recommendations & 0.30 & -0.14 & 0.26 & -0.16 & -0.38 & \textbf{0.54}\\
Policy Limits & 0.17 & -0.05 & 0.08 & 0.02 & -0.11 & \textbf{0.68} \\
\bottomrule
\end{tabular}
\caption{Pearson correlation coefficients of different indicators to answer correctness, grouped by tasks.
The largest absolute correlation for each task type is bolded.}
\label{correlation}
\end{table*}

Our analysis in the main paper presented correlations and a lack thereof between various indicators and answer correctness over all traces.
We additionally performed a correlation analysis per task type.
Table \ref{correlation} reports measured Pearson correlation coefficients between answer correctness on traces and 6 indicators grouped by 6 non-overlapping task types.
As reported in the main paper, tool errors have little or weak observed correlation with answer correctness.
However, we do see more significant correlations with tool error recovery and answer correctness.
Here, tool error recovery is the rate of self-corrected tool calls, i.e., if a tool error is made, how often a corrected call to the same tool is subsequently made.

\subsection{Answer Correctness pass\textasciicircum k by Task Type}
\label{app:}

\begin{figure*}[ht]
    \centering
    \includegraphics[width=0.8\textwidth]{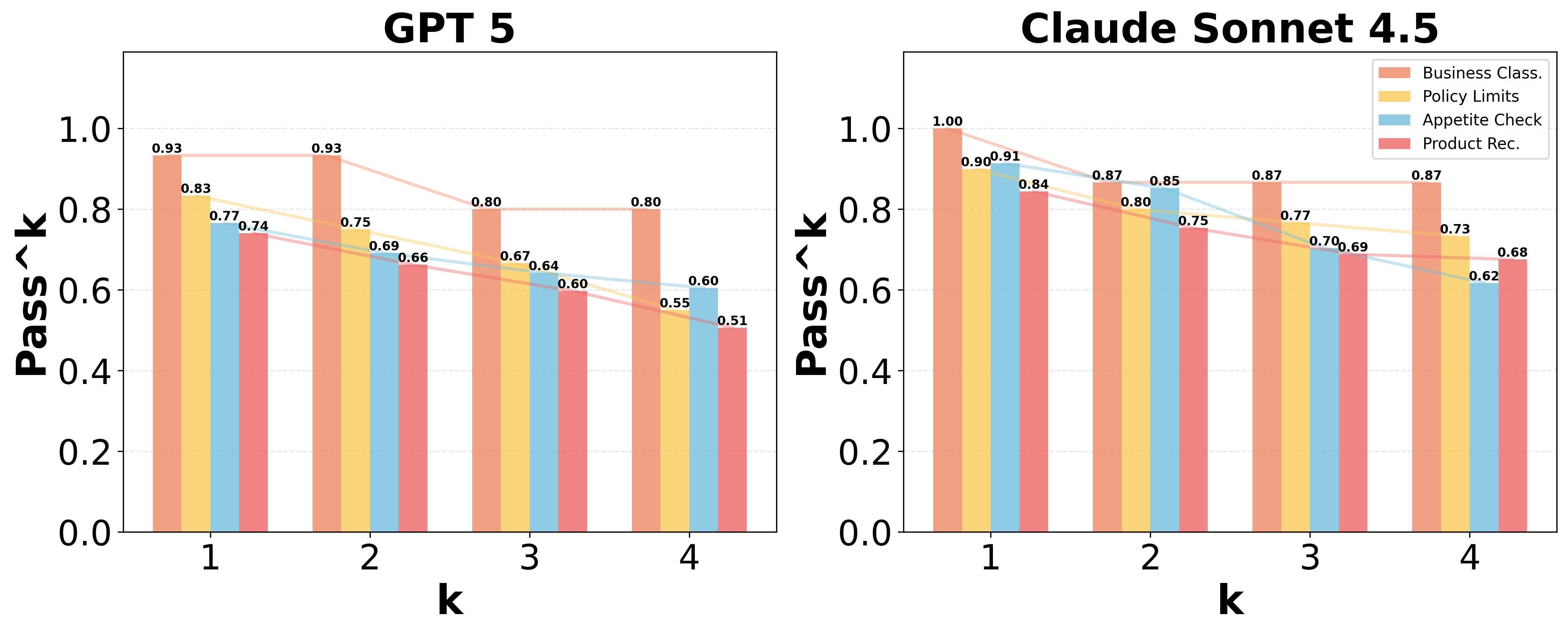}
    \caption{Answer correctness pass\textasciicircum k for GPT-5 and Claude Sonnet 4.5 across a selected subset of task types.}
    \label{fig:pass_k_by_task_type}
\end{figure*}

The main paper reported overall pass\textasciicircum k metrics for GPT-5 and Claude Sonnet 4.5, the 2 models with the highest answer correctness.
This metric is important for measuring reliability and has real implications on the applicability of agents.
Figure \ref{fig:pass_k_by_task_type} shows our analysis of pass\textasciicircum k metrics for GPT-5 and Claude Sonnet 4.5 grouped by 4 selected task types.
The pass\textasciicircum k results show slightly different trends for each model and task type.
Overall, Claude Sonnet 4.5 maintains a higher rate of answer correctness. 
Moreover, Figure \ref{fig:pass_k_by_task_type} shows a roughly $20$ point decrease from  pass\textasciicircum 1 to pass\textasciicircum 4 for Product Recommendation in both models.
This type of specialized error mode analysis is useful for target application and training.